%% file: arxiv-submission.tex
\documentclass[letterpaper]{article} 
\usepackage[]{aaai2026}  
\usepackage{times}  
\usepackage{helvet}  
\usepackage{courier}  
\usepackage[hyphens]{url}  
\usepackage{graphicx} 
\usepackage{natbib}  
\usepackage{caption} 
\usepackage{subcaption}
\usepackage{algorithm}
\usepackage{algorithmic}
\usepackage{eurosym}
\usepackage{enumitem}
\usepackage{amsmath, amssymb}
\usepackage{multirow}
\usepackage[table,xcdraw]{xcolor}
\usepackage{tabularx}

\usepackage{booktabs}

\usepackage{newfloat}
\usepackage{listings}
\DeclareCaptionStyle{ruled}{labelfont=normalfont,labelsep=colon,strut=off} 
\floatstyle{ruled}
\newfloat{listing}{tb}{lst}{}
\floatname{listing}{Listing}
\nocopyright

\title{RainSeer: Fine-Grained Rainfall Reconstruction via Physics-Guided Modeling}
\author {
    Lin CHEN\textsuperscript{\rm 1},
    Jun CHEN\textsuperscript{\rm 1},
    Minghui QIU\textsuperscript{\rm 1},
    Shuxin ZHONG\textsuperscript{\rm 1},
    Binghong CHEN\textsuperscript{\rm 2},
    Kaishun WU\textsuperscript{\rm 1}
}
\affiliations {
    \textsuperscript{\rm 1}The Hong Kong University of Science and Technology (Guangzhou)\\
    \textsuperscript{\rm 2}China Meteorological Administration\\
    lchen297@connect.hkust-gz.edu.cn, jchen512@connect.hkust-gz.edu.cn, mqiu585@connect.hkust-gz.edu.cn, shuxinzhong@hkust-gz.edu.cn, chenbinghong@163.com, wuks@hkust-gz.edu.cn
}

\usepackage{bibentry}
\input{9cmd}

\begin{document}

\maketitle

\input{0abstract}
\input{1intro}
\input{2relatedwork}

\input{3preliminaries}

\input{4framework3_cj}
\input{5experiments}

\input{7conclusion}

\bibliography{aaai2026}

\end{document}

%% file: 9cmd.tex
\newcommand{\N}{\texttt{RainSeer}}

\newcommand{\ComponentA}{Structure-to-Point Mapper}
\newcommand{\ComponentB}{Geo-Aware Rain Decoder}

\newcommand{\AsubComponentA}{Hierarchical Multi-Modal Encoders}
\newcommand{\AsubComponentB}{Cross-Scale Aligner}

%% file: 0abstract.tex
\begin{abstract}
Reconstructing high-resolution rainfall fields is essential for flood forecasting, hydrological modeling, and climate analysis.
However, existing spatial interpolation methods—whether based on automatic weather station (AWS) measurements or enhanced with satellite/radar observations-often over-smooth critical structures, failing to capture sharp transitions and localized extremes.
We introduce~\N, a structure-aware reconstruction framework that reinterprets radar reflectivity as \textbf{a physically grounded structural prior}—capturing when, where, and how rain develops.
This shift, however, introduces two fundamental challenges: 
(i) translating high-resolution volumetric radar fields into sparse point-wise rainfall observations,
and (ii) bridging the physical disconnect between aloft hydro-meteors and ground-level precipitation. 
\N\ addresses these through a physics-informed two-stage architecture:  
a \textit{\ComponentA} performs spatial alignment by projecting mesoscale radar structures into localized ground-level rainfall, through a \textit{bidirectional mapping}, 
and a \textit{\ComponentB} captures the semantic transformation of hydro-meteors through descent, melting, and evaporation via a \textit{causal spatiotemporal attention} mechanism.  
We evaluate \N\ on two public datasets—RAIN-F (Korea, 2017–2019) and MeteoNet (France, 2016–2018)—and observe consistent improvements over state-of-the-art baselines, reducing MAE by over 13.31\% and significantly enhancing structural fidelity in reconstructed rainfall fields.
\end{abstract}

%% file: 1intro.tex
\section{Introduction}
\label{sec:intro}

Rainfall is far from uniform—it clusters in some regions and vanishes in others~\cite{westra2014future}. 
However, sparse AWS networks may fail to capture this fine-grained variability~\cite{hrachowitz2011uncertainty},
providing hydrological models with incomplete or biased inputs~\cite{devia2015review}.
The consequences are tangible:
during the 2021 Eifel floods in Germany~\cite{Europeanfloods}, mislocated storm centers due to sensor distortion delayed alerts, contributing to 243 deaths and over \euro{}46 billion in damages.
Such failures reveal that rainfall field reconstruction is not just a technical objective but also a critical safeguard for disaster resilience and human survival.

\begin{figure}
\centering
\includegraphics[width=1\linewidth]{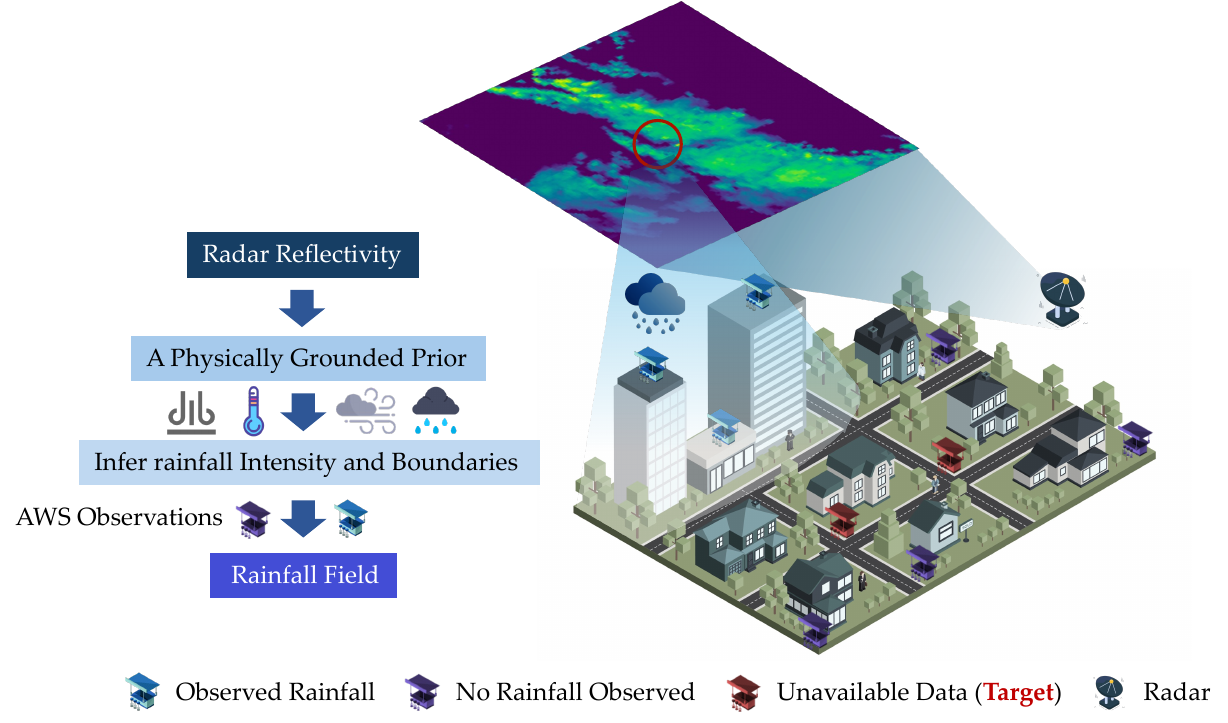}
\caption{
Modeling Strategy of Abrupt Rainfall Field.
}
\label{fig:scenario}
\end{figure} 

Existing approaches rely on two primary sources.  
(1) \textit{Ground-based methods} (e.g., kriging~\cite{lucas2022optimizing}, inverse distance weighting~\cite{tan2021coupling}, and splines~\cite{vimelia2024prediction}) interpolate sparse AWS data under spatial smoothness assumptions. 
(2) \textit{Remotely sensed methods} (e.g., satellite- and radar-driven estimates~\cite{abdollahipour2022review, sokol2021role}) provide wider coverage but often map precipitation fields as spatially uniform. 
Both paradigms struggle to resolve the very features that matter most—\textit{localized extremes and abrupt transitions}—undermining reliability in high-stakes scenarios~\cite{xu2022era5, li2023ssin}.

Radar is more than a snapshot of where it rains—it encodes when, where, and how storms unfold~\cite{wu2000dynamical}. 
Deep echo tops hint at updraft-driven hail; 
bow echoes trace damaging winds; 
hook echoes expose rotating mesocyclones.
Even as convection fades, stratiform shields still shape surface rainfall.
These patterns reflect the \textit{physics of storms}, not just their appearance.  
This opens a pivotal opportunity:  
can we move beyond signal matching and reconstruct rainfall by reasoning from storm physics, treating radar as \textit{a physically grounded, evolving prior}?
 
Despite straightforward, this idea involves two fundamental challenges. 
First, radar provides high-resolution volumetric scans that capture mesoscale convective structures, whereas station measurements are sparse and point-based, creating a scale and sampling gap (\textbf{C1}).
Second, radar reflectivity senses hydro-meteors aloft, yet what reaches the ground is shaped by descent, melting, and evaporation.
This leads to \textit{temporal lags, spatial shifts, and intensity inconsistencies} between what radar sees and what actually rains (\textbf{C2}).
See the \textit{Preliminary} section for details.

Thus, we design~\N, a structure-aware fine-grained rainfall reconstruction framework that treats radar reflectivity as a physically grounded, evolving prior. 
To address \textbf{C1}, \N\ introduces \textit{\ComponentA} performs physically guided spatial alignment:  
i) \textit{\AsubComponentA}, which includes a~\textit{Radar Encoder} distills mesoscale storm structures—and a~\textit{AWS Encoder} propagates rainfall-relevant signals across the AWS network;
and ii) a \textit{\AsubComponentB}, which leverages \textit{Bidirectional Cross Attention} to align them, mapping radar signals into localized rainfall footprints. 
To address \textbf{C2},~\N\ introduces a~\textit{\ComponentB}, equipped with a \textit{Causal Spatiotemporal Attention (CSTA)} mechanism that learns how hydro-meteors evolve downward, capturing their decay through transport, melting, and evaporation.
CSTA serves as a learnable structural prior, correcting semantic gaps, and enforcing physical consistency in rainfall reconstruction.
Our main contributions are listed as follows:
\begin{itemize} [leftmargin=*]
    \item 
    We introduce a new paradigm that treats radar reflectivity as \textit{a physically grounded structural prior} for rainfall field reconstruction, enabling finer recovery of spatial discontinuities and localized extremes.
    
    \item \N\ designs two components:
    the~\textit{\ComponentA} performs spatial alignment, projecting mesoscale radar structures onto ground-level rainfall through a \textit{bidirectional mapping},  
    and the~\textit{\ComponentB} models the semantic transformation of hydro-meteors during their descent, using a \textit{causal spatiotemporal attention} mechanism to capture melting, evaporation, and transport dynamics—serving as a learnable structural prior that enforces physical consistency and temporal coherence.

    \item We evaluate \N\ using two public datasets RAIN-F (Korean, 2017-2019) and MeteoNet (France, 2016-2018). 
    \N\ consistently outperforms all baselines in both MeteoNet and RAIN-F, reduced RMSE by 9.28\% and 47.37\%, MAE by 13.31\% and 70.71\%, while boosting NSE by 60.75\% and 22.56\%, and CC by 11.15\% and 10.48\%, respectively. 
\end{itemize}

%% file: 2relatedwork.tex
\section{Related Work}
\label{sec:related_work}


\subsection{Rainfall Spatial Interpolation}
Spatial interpolation of rainfall has long been a central task in environmental science and hydrology~\cite{prein2017increased}.
Traditional rainfall interpolation methods can be broadly divided into single-source and multi-source approaches.
\textit{Single-source methods}, which rely solely on gauge observations, are typically categorized into two families~\cite{ly2013different}:
i) \textit{deterministic approaches} (e.g., Inverse Distance Weighting~\cite{lu2008adaptive}, Thin Plate Splines~\cite{hutchinson1995interpolating}, and TIN~\cite{song2021efficient}), which rely on geometric rules without modeling uncertainty, 
and ii) \textit{geostatistical methods} (e.g., Ordinary Kriging~\cite{wackernagel2003ordinary}), which model spatial correlation through statistical variograms. Both types typically assume spatial stationarity and perform best in smooth terrains.
To enhance spatial coverage and resolution, multi-source frameworks have emerged that integrate satellite or radar data with ground stations~\cite{abdollahipour2022review, sokol2021role}.
However, these methods often employ empirical mappings that implicitly assume precipitation is spatially homogeneous and statistically stationary—assumptions that fail precisely where high-resolution accuracy is most critical, such as near convective cores or sharp frontal boundaries~\cite{li2023ssin}.

\subsection{Spatiotemporal Learning}
Recent advances in AI and machine learning have revitalized rainfall estimation by enabling more flexible modeling of spatial and temporal dependencies.
CNNs~\cite{zhang2021deep, guo2019attention} excel at extracting local features from gridded inputs like radar or satellite data;
GNNs~\cite{ijcai2024p269, ijcai2024p803} generalize spatial modeling to non-Euclidean domains;
and Transformers~\cite{li2023ssin, liang2023airformer} offer a unified framework for capturing long-range spatiotemporal dependencies via self-attention.

\subsection{Causal Inference}
Causal inference offers an interpretable and robust modeling paradigm by capturing stable, invariant relationships that hold under varying conditions, beyond mere correlations~\cite{pearl2009causal, pearl2009causality}, which are particularly valuable for precipitation estimation~\cite{wu2024earthfarsser, wang2024nuwadynamics}, where physical drivers (e.g., uplift, moisture transport, and microphysical processes) and their interactions vary markedly across space, time, and weather regimes.

%% file: 3preliminaries.tex
\section{Preliminary}
\label{sec:preliminary}



\subsection{Motivation}

Rainfall is rarely smooth, stationary, or predictable. 
It flares up, shifts rapidly, and fades just as fast—especially during convective storms.  
Yet most reconstruction models rely on extrapolating point observations under assumptions of continuity, locality, and temporal persistence. 
These work for gentle drizzles, but break down when it matters most: at the onset of a convective surge, along a frontal collision zone, or over terrain-forced rain bands.

Weather radar offers a fundamentally different lens. 
It does not just show where it is raining—it captures how storms are built.  
A towering echo top rising to 15 km may indicate an updraft about to unload hail.  
A bow echo curving through a squall line might precede damaging straight-line winds.  
A hook echo coiled within a supercell can signal mesocyclone rotation and downburst potential.  
Even after the main convection subsides, a trailing stratiform shield often lingers and contributes to cumulative rainfall.  
Radar reveals these structures not as isolated events, but as evolving spatial patterns—storm morphology unfolding in real time.
These radar signatures are tightly linked to surface rainfall outcomes.  
They trace when rain is likely to form, where it will concentrate, and how long it may persist.  
In this sense, radar reflectivity is more than just input—it is a \textbf{dynamic, physically structured prior}.

\subsection{Challenges}
    

At first glance, the idea seems intuitive: radar sees storms, stations record rain—why not simply connect the two?
Yet beneath this apparent simplicity lie two key challenges.
\begin{itemize} [leftmargin=*]
    \item \textbf{(C1) Spatial Resolution Mismatch—Structured Fields vs. Sparse Points.}
    Radar offers sweeping, high-resolution volumetric scans that reveal the spatial anatomy of storms—from vertically developed convective towers to elongated frontal bands and broad stratiform shields.
    These 3D fields resolve storm morphology at a kilometer-scale granularity across entire regions.
    In contrast, AWS stations are sparse and static, providing isolated, point-based measurements at the surface.
    Bridging these two domains—dense, structured radar fields and sparse, terrain-sensitive station readings—is inherently ill-posed, especially in complex environments where rainfall is localized, topographically influenced, or driven by sub-kilometer dynamics invisible to the gauge network.
    \item \textbf{(C2) Semantic Misalignment—What Radar Sees is not always What Falls.}
    Radar reflectivity reveals hydrometeors suspended aloft—but AWSs only record what actually reaches the ground.
    Between those layers lies a cascade of microphysical processes: particles may descend, melt, and sometimes evaporate mid-air.
    This incurs semantic gaps:
    (i) \textit{Temporal lag}—echo growth aloft often precedes surface rainfall by several minutes;
    (ii) \textit{Spatial shift}—tilted storm structures and wind advection push rainfall away from radar-indicated cores;
    and (iii) \textit{Intensity ambiguity}—a given reflectivity may signify anything from virga to hail to torrential rain, depending on sub-cloud humidity, thermodynamic stratification, and terrain interactions.
\end{itemize}

\subsection{Problem Formulation}
We formulate high-resolution rainfall field reconstruction as a spatio-temporal estimation task guided by radar-derived structural priors.
Let $\mathcal{S}_t = {(x_i, y_i, p_i)}_{i=1}^{N}$ denote sparse surface rainfall observations at time $t$, where $(x_i, y_i)$ is the location of the AWS $i$ and $p_i$ is the recorded rainfall intensity. 
Similarly, let $\mathcal{R}_t = {(x_j, y_j, z_j)}_{j=1}^{M}$ represent the radar reflectivity field, where $z_j$ is the reflectivity value on grid $j$ and $M$ is the number of grids.
Note that $M \gg N$, reflecting the significantly higher spatial resolution of radar.
Our goal is to learn a reconstruction model $\mathcal{F}_\theta$ that reconstructs the dense rainfall field $\hat{\mathcal{C}}_{T}$, conditioned on multi-step sparse observations $\mathcal{S}_{[t_0:t_0+T]}$ and radar reflectivity sequences $\mathcal{R}_{[t_0:t_0+T]}$: 
\begin{equation}
    \hat{\mathcal{C}_{T}} = \mathcal{F}_\theta(\mathcal{S}_{[t_0:t_0+T]}, \mathcal{R}_{[t_0:t_0+T]}).
    \label{eq:formulation}
\end{equation}

%% file: 4framework3_cj.tex
\section{Methodology}
\label{sec:method}

\begin{figure*}[t]
  \centering
  \includegraphics[width=\textwidth]{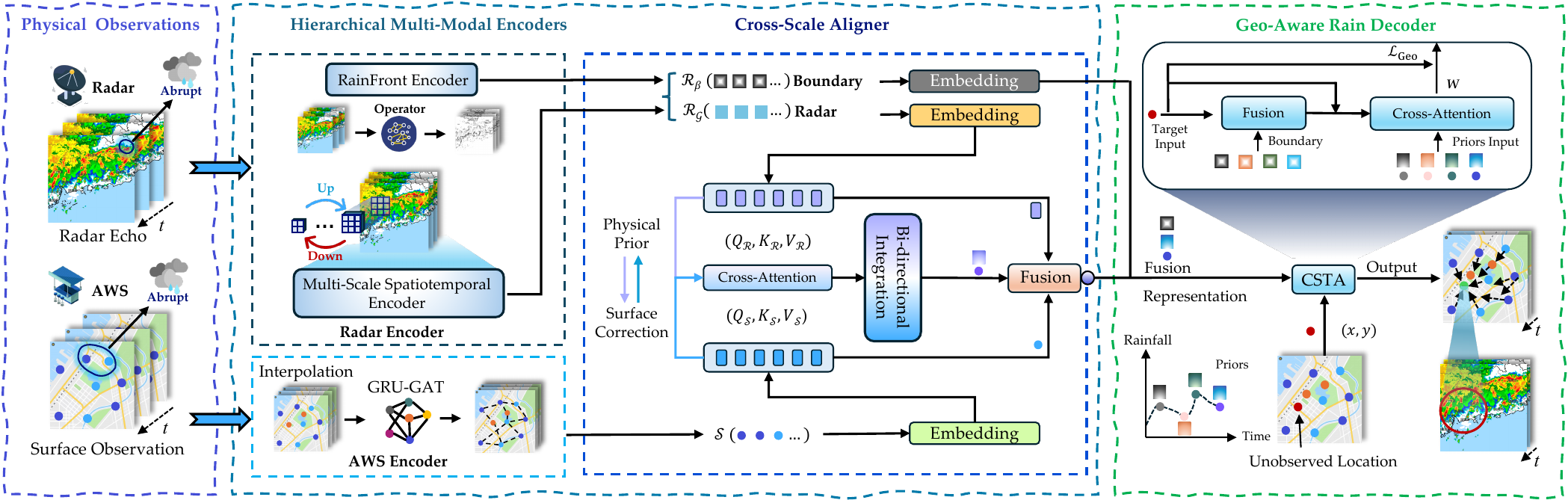}
  \caption{The Framework of~\N.}
  \label{fig:framework}
\end{figure*}

To reconstruct high-fidelity rainfall fields, we introduce \N, a structure-aware framework composed of two physically grounded components (as shown in Figure~\ref{fig:framework}):
\begin{itemize} [leftmargin=*]
    \item \textbf{\ComponentA} bridges the spatial scale gap by learning how volumetric storm structures aloft (e.g., convective towers, frontal bands) translate into localized surface rainfall.
    It first employs ~\textit{\AsubComponentA}: a \textit{Radar Encoder} captures physically salient signatures, while a \textit{AWS Encoder} propagates rainfall-relevant signals across the AWS network.
    Then, a~\textit{\AsubComponentB} designs \textit{Bidirectional Physics-Aware Attention} to reconcile radar volumes with point-level gauges, guided by spatial priors (e.g., distances).

    \item \textbf{\ComponentB} targets the semantic gap—when radar sees clouds but not rain—by modeling the downstream microphysics (e.g., descent, melting, evaporation) that modulate what reaches the ground.
    It designs a~\textit{Causal Spatiotemporal Attention (CSTA)} mechanism that acts as a \textit{learnable structural prior}, ensuring the reconstructed rainfall respects storm morphology, honors physical boundaries, and evolves plausibly over time.
\end{itemize}

\subsection{\ComponentA}
\label{subsec:componentA}
\textit{\ComponentA} addresses the scale and modality gap between 3D radar volumes and sparse station observations via two physically grounded modules:~\textit{\AsubComponentA} and \textit{\AsubComponentB}.

\subsubsection{\AsubComponentA} extract physically grounded, modality-specific representations.

\textbf{Radar Encoder} is designed to extract both coherent mesoscale patterns and fine-grained physical discontinuities.

\begin{itemize} [leftmargin=*]

   \item \textbf{Multi-Scale Spatiotemporal Encoder}
   To jointly capture short-term temporal dependencies and local spatial structures, we first apply a spatial-temporal separable convolution (STSC) to the radar reflectivity sequence $\mathcal{R}_{[t_0:t_0+T]}$, where temporal and spatial features are extracted via a sequence of 2D spatial and 1D temporal convolutions:
    \begin{equation}
        \mathbf{E}_{\text{STSC}} = \mathrm{\text{STSC}}(\mathcal{R}_{[t_0:t_0+T]}).
    \end{equation}
    To enrich multi-scale spatial context, we adopt a multi-scale residual inception~\cite{Szegedy_2015_CVPR} with three parallel branches operating at progressively downsampled resolutions (full, $1/2$, and $1/4$).
    Each lower-resolution feature map is upsampled and fused with the main path:
    \begin{align}
    \ \widetilde{\mathbf{E}}_{\text{STSC}} &= \text{Inception}(\mathbf{E}_{\text{STSC}}), \nonumber \\
    \mathbf{E}_{1/2} &= \text{Inception}(\text{Down}_{2}(\mathbf{E}_{\text{STSC}})), \nonumber \\
    \mathbf{E}_{1/4} &= \text{Inception}(\text{Down}_{4}(\mathbf{E}_{\text{STSC}})), \nonumber \\
    \mathbf{E}_{\text{radar}} &= \phi\left( \widetilde{\mathbf{E}}_{\text{STSC}}     + \text{Up}_{2}(\mathbf{E}_{1/2}) + \text{Up}_{4}(\mathbf{E}_{1/4}) \right),
    \end{align}
    where $\phi(\cdot)$ denotes post-fusion transformation.
    To encode spatiotemporal positional information, we add two learnable embeddings:
    a temporal embedding $e_t^{\text{time}}$ indicating the current time step $t$, and a spatial embedding $e_s^{\text{space}}$ representing the relative position of each grid, defined as:
        \begin{equation}
        \mathcal{H}_{\text{radar}} = \mathbf{E}_{\text{radar}}  + e_t^{\text{time}} + e_s^{\text{space}}.
    \end{equation}

    \item \textbf{RainFront Encoder}
    To highlight motion discontinuities and structural transitions, we extract boundary priors from radar reflectivity via a Laplacian operator $\Delta$, computed as:
    \begin{equation}
        \mathbf{B}_t = \Delta * \mathcal{R}_t, \quad t = 1, \dots, T,
    \end{equation}
    where $*$ denotes convolution. The resulting boundary sequence $\mathbf{B}_{1:T}$ is then passed into a ConvLSTM to model temporal boundary dynamics:
    \begin{equation}
        \mathcal{H}_{\text{boundary}} = \mathrm{ConvLSTM}(\mathbf{B}_t).
    \end{equation}
\end{itemize}


\textbf{AWS Encoder} models ground stations as nodes in a graph $\mathcal{G} = \{\mathcal{V}, \mathcal{E}\}$, with edges encode proximity-based atmospheric correlations.
To capture more spatial correlation information, we first use radar data for random interpolation to create more virtual nodes:
\begin{equation}
  \hat{g}_{i,t}  = \text{Interp}(\mathcal{R}_{\text{radar}}).
\end{equation}
At each time step $t$, the input features of each node are raw gauge observation, spatial and temporal embeddings:
\begin{equation}
    h^{(l)}_{i,t} = ReLU \left( W_{\text{in}} [g_{i,t} ; e^{\text{time}}_t ; e^{\text{space}}_i] \right),
\end{equation}
where $g_{i,t}$ is the observation of node $i$ at time $t$, $[;]$ is concatenation.
To capture spatiotemporal dependencies, the encoder stacks $L$ layers, each combining a graph attention network (GAT)~\cite{GAT} for spatial message passing and a gated recurrent unit (GRU) cell for temporal propagation:
\begin{equation}
    \begin{aligned}
    z^{(l)}_{i,t} &= \text{ELU}\left(\text{GAT}^{(l)}\left(h^{(l)}_{i,t}, \{h^{(l)}_{j,t} \mid j \in \mathcal{N}(i)\}\right)\right), \\
h^{(l)}_{i,t} &= \text{GRU}^{(l)}\left(z^{(l)}_{i,t}, h^{(l)}_{i,t-1}\right),
    \end{aligned}
\end{equation}
where ELU denotes an exponential linear unit activation function \cite{clevert2016fastaccuratedeepnetwork} and $h^{(l)}_{i,t}$ is the hidden state of node $i$ at layer $l$ and time $t$.
The resulting embeddings $\mathcal{H}_{\text{aws}} = \{ h^{(L)}_{i,t} \mid i \in \mathcal{V}, t = 1, \dots, T \}$ encode evolving rainfall patterns and regional coherence, guiding high-resolution reconstruction.

\subsubsection{\AsubComponentB} bridges the modality gap between grid-based radar field and graph-based AWS observations.
Specifically:
i) Radar$\rightarrow$AWS attention reflects how storm structures aloft influence surface precipitation;
and ii) AWS$\rightarrow$Radar attention incorporates ground-truth feedback to refine radar representations.
Given radar features $\mathcal{H}_{radar}$, we split each frame into non‑overlapping $p\times p$ patches:
\begin{equation}
    \hat{\mathcal{H}}_{radar}=\mathrm{Proj}_r\big(\mathrm{Patchify}(\mathcal{H}_{radar},p)\big).
\end{equation}
AWS features $\mathcal{H}_{\text{aws}}$ are flattened over time and stations and projected to the same latent size:
\begin{equation}
    \hat{\mathcal{H}}_{\text{aws}}=\mathrm{Proj}_s\big(\mathrm{Flatten}_{T,N}(\mathbf{\mathcal{H}_{\text{aws}}})\big).
\end{equation}
We then design a~\textit{Bidirectional Physics-Aware Multi-Head Attention (BPA-MHA)} mechanism to model the causal interplay between radar signals aloft and surface rainfall observations.

\begin{align}
    \mathbf{E}_{R\leftarrow S}&=\mathrm{MHA}\!\left(Q=\hat{\mathcal{H}}_{radar},\,K=\hat{\mathcal{H}}_{\text{aws}},\,V=\hat{\mathcal{H}}_{\text{aws}}\right), \nonumber \\
    \mathbf{E}_{S\leftarrow R}&=\mathrm{MHA}\!\left(Q=\hat{\mathcal{H}}_{\text{aws}},\,K=\hat{\mathcal{H}}_{radar},\,V=\hat{\mathcal{H}}_{radar}\right).
\end{align}
For each modality, we concatenate the attended features with the original tokens along the feature dimension:
\begin{align}
    \tilde{\mathbf{R}}&=\big[\mathbf{E}_{R\leftarrow S}\,\|\,\hat{\mathcal{H}}_{radar}\big], \nonumber \\
    \tilde{\mathbf{S}}&=\big[\mathbf{E}_{S\leftarrow R}\,\|\,\hat{\mathcal{H}}_{\text{aws}}\big].
\end{align}
Finally, we concatenate $\tilde{\mathbf{R}}$ and $\tilde{\mathbf{S}}$ along the sequence dimension to form the output memory:
\begin{equation}
    \mathcal{H}_{\text{aligned}}=\mathrm{Concat}(\tilde{\mathbf{R}},\tilde{\mathbf{S}}).
\end{equation}


\subsection{\ComponentB}
\label{subsec:componentB}
\textit{\ComponentB} addresses the semantic gap—where radar sees cloud structures aloft but not all lead to surface rainfall—by explicitly modeling the downstream microphysical processes (e.g., descent, melting, evaporation) that govern precipitation formation.

To enable efficient and spatially controllable rainfall recovery, we adopt a query-based decoding mechanism that reconstructs rainfall at any specified spatiotemporal location.
Given a query position $(x_q, y_q)$, we first encode it into a latent query representation:
\begin{equation}
\mathbf{q}_{loc} = \text{MLP}(x_q, y_q)
\end{equation}
This representation is then formed by combining a spatio-temporal embedding of the query location with boundary information of the radar, through a fusion module to integrate boundary features into the initial query embedding, enriching the query with information from the surrounding:
\begin{equation}
\mathbf{q}_q = \text{MLP}(\mathbf{q}_{loc} + \text{Fuse}(\mathbf{q}_{loc}, \mathcal{H}_{\text{boundary}}))
\end{equation}
The enriched query $\mathbf{q}_q$ is then used to retrieve relevant information from a historical context set, $\mathcal{H}_{\text{aligned}}$, which contains latent features from radar and AWS observations.

However, this context may contain both (i) causal priors $\mathbf{c}_i$ that physically influence the target rainfall (e.g., upstream reflectivity cores, persistent frontal zones), and (ii) confounding patterns that are statistically correlated but not causally linked (e.g., remote storms, decayed convection).
Thus, we filter historical signals to isolate only those with genuine causal influence, motivated by causal theory~\cite{pearl2009causality,wang2024nuwadynamics}.
The implemented model implicitly learns this relationship through \textit{CSTA}. Specifically, we compute causal attention weights using both content relevance and geo-spatiotemporal proximity:
\begin{equation}
    \alpha_i = \frac{ \exp\left( \mathbf{q}_q^\top \mathbf{W_a} \mathbf{k}_i \right) }{ \sum_{j: t_j \le t_q} \exp\left( \mathbf{q}_q^\top \mathbf{W_a} \mathbf{k}_j\right) },
\end{equation}
where $\mathbf{W_a}$ is a project of learnable attention, $\mathbf{k}_i$ is a latent key from the historical context set $\mathcal{H}_{\text{aligned}}$, and $t_q$ is the timestamp of the query.
The retrieved causal prior is then:
\begin{equation}
    \mathbf{c}_{\text{prior}} = \sum_{i: t_i \le t_q} \alpha_i \cdot \mathbf{v}_i,
\end{equation}
where $\mathbf{v}_i$ is the value vector corresponding to the key $\mathbf{k}_i$.
The final rainfall prediction at the query location is obtained:
\begin{equation}
    \hat{r}(x_q, y_q) = \psi(\mathbf{c}_{\text{prior}}, \mathbf{q}_q),
\end{equation}
where $\psi(\cdot)$ is a learnable predictor.

\subsection{Training Objective}

The model is trained to predict rainfall at arbitrary geographic locations by minimizing a combination of a standard mean squared error (MSE) loss and a geographically-informed regularization term, \textbf{GeoLoss}.

The MSE loss ensures accurate supervision at labeled locations, while \textbf{GeoLoss} encourages spatial consistency by aligning attention vectors of nearby locations—following Tobler’s First Law~\cite{miller2004tobler}.
Formally, it is defined as:
\begin{equation}
    \mathcal{L}_{\text{Geo}} = \frac{1}{|P|} \sum_{(i,j) \in P} \text{Sim}(\mathbf{p}_i, \mathbf{p}_j) \cdot \left(1 - \text{Sim}(\mathbf{a}_i, \mathbf{a}_j) \right),
\end{equation}
where $\mathbf{p}_i$ and $\mathbf{a}_i$ denote the position embedding and normalized attention vector for query location $i$, respectively, $\text{Sim}(\cdot, \cdot)$ is the cosine similarity between position embeddings or attention vectors, $P$ is the set of all distinct query pairs.
The overall training objective is:
\begin{equation}
    \mathcal{L} = \mathcal{L}_{\text{MSE}} + \lambda \cdot \mathcal{L}_{\text{Geo}},
\end{equation}
where $\lambda$ balances prediction accuracy and spatial regularity.



%% file: 5experiments.tex
\section{Evaluation}
\label{sec:evaluation}

This section presents a comprehensive evaluation of our~\N\ by addressing the following key questions:
\begin{itemize} [leftmargin=*]
    \item~\textbf{RQ1}: What is \N's overall performance?
    \item~\textbf{RQ2}: How does each component affect performance?
    \item~\textbf{RQ3}: What is the effect of the major hyperparameters?
\end{itemize}

\begin{table*}[htbp]
\centering
{
\begin{tabular}{lcccccccc}
\toprule
\multirow{2}{*}{Model} & \multicolumn{4}{c}{\textbf{MeteoNet Dataset}} & \multicolumn{4}{c}{\textbf{RAIN-F Dataset}} \\
\cmidrule(lr){2-5} \cmidrule(lr){6-9}
& RMSE↓ & MAE↓ & NSE↑ & CC↑ & RMSE↓ & MAE↓ & NSE↑ & CC↑ \\
\midrule
TIN         & 0.865 & \underline{0.081} & -0.182 & 0.255 & 0.636 & \underline{0.086} & 0.739 & 0.860 \\
TPS         & 0.997 & 0.125 & -0.517 & 0.216 & \underline{0.608} & 0.093 & \underline{0.762} & \underline{0.875} \\

ZR          & \underline{0.713} & 0.098 & \underline{0.226}  & 0.484 & 1.015 & 0.209 & 0.337 & 0.595 \\
QPENet      & 0.778 & 0.091 & 0.078  & \underline{0.545} & 1.056 & 0.178 & 0.283 & 0.620 \\

SSIN        & 0.813 & 0.124 & -0.007 & 0.170 & 0.715 & 0.104 & 0.671 & 0.834 \\
GSI         & 0.853 & 0.105 & -0.109 & 0.261 & 0.622 & 0.100 & 0.751 & 0.867 \\
STFNN       & 0.828 & 0.337 & 0.029  & 0.257 & 0.661 & 0.135 & 0.718 & 0.848 \\

\textbf{RainSeer} & \textbf{0.647} & \textbf{0.070} & \textbf{0.363} & \textbf{0.605} & \textbf{0.320} & \textbf{0.025} & \textbf{0.934} & \textbf{0.967} \\
\bottomrule
\end{tabular}
}
\caption{Quantitative comparison on MeteoNet and RAIN-F datasets. Best results are in \textbf{bold}, second best are \underline{underlined}.}
\label{tab:overall}
\end{table*}

\subsection{Evaluation Settings}
\subsubsection{Datasets.}
We evaluate \N\ on two public rainfall datasets:
i) \textbf{RAIN-F}~\cite{RAIN-F}, which provides hourly radar and AWS measurements from 900 stations in South Korea (2017–2019),
and ii) \textbf{MeteoNet}~\cite{MeteoNet}, which offers radar measurements at 5-minute intervals and AWS measurements at 6-minute intervals from 287 stations across France (2016–2018).
All data is resampled to 1-hour resolution for consistency.

\subsubsection{Baseline Methods.}
We compare our~\N\ with strong baselines that belong to 3 categories:
\begin{itemize} [leftmargin=*]
    \item\textbf{Statistical models}: \textbf{TPS}~\cite{hutchinson1995interpolating} and \textbf{TIN}~\cite{lu2008adaptive} are widely used in geoscience and hydrology for rainfall field reconstruction from AWS.
    \item\textbf{Radar-based models}: \textbf{Z-R}~\cite{peng2022radar} uses conventional reflectivity–rainrate conversion, while \textbf{QPENet}~\cite{zhang2021deep} is a CNN-based method trained to map reflectivity to precipitation.
    In the Z-R method, the parameters a and b are set to 200 and 1.6, respectively, as this set of parameters achieves the best performance across both datasets.
    These models benchmark the efficacy of radar-only estimation.
    \item\textbf{Spatiotemporal deep learning models}: We include representative spatiotemporal models: \textbf{SSIN}~\cite{li2023ssin}, \textbf{GSI}~\cite{li2023rainfall}, and \textbf{STFNN}~\cite{ijcai2024p803}—all designed for station-level extrapolation tasks such as rainfall or air pollution field prediction.
\end{itemize}

\subsubsection{Metrics.}
To evaluate the performance of \N, we adopt four widely used metrics:
MAE, RMSE, Nash–Sutcliffe Efficiency (NSE) \cite{li2023ssin}, and Correlation Coefficient (CC) \cite{li2023polarimetric}.
Together, these metrics capture intensity accuracy and spatialtemporal consistency, critical aspects of rainfall forecasting.

Specifically, NSE reflects the proportion of variance explained—values close to 1 indicate strong predictive power relative to the climatological mean:
\begin{equation}
\text{NSE} = 1 - \frac{\sum_{i=1}^N (G_i - R_i)^2}{\sum_{i=1}^N (G_i - \bar{G})^2}
\end{equation}
where $G_i$ and $R_i$ are the ground truth and prediction at sample $i$, $N$ the number of samples, and $\bar{G}$ the mean ground truth.
CC measures linear correlation and spatial alignment:
\begin{equation}
    \text{CC} = \frac{\sum_{i=1}^N (G_i - \bar{G}) \cdot (R_i - \bar{R})}{\sqrt{\sum_{i=1}^N (G_i - \bar{G})^2} \cdot \sqrt{\sum_{i=1}^N (R_i - \bar{R})^2}}
\end{equation}
where \( \bar{R} \) is the mean of predictions.

\subsubsection{Implementation Details.}
\N\ is implmented using PyTorch and trained on a single NVIDIA A800 GPU and a Intel(R) Xeon(R) Platinum 8358P CPU @ 2.60GHz for 36,000 steps with batch size 32.
For optimization, we use AdamW with an initial learning rate of $5 \times 10^{-4}$, scheduled by a OneCycle policy with the same peak.
GeoLoss weight ($\lambda$) is set to 0.1 to achieve optimal performance.
The number of nearest neighbor nodes, k, in the AWS Encoder is set to 8.
The time window is configured to 60 minutes.
For evaluation, 20\% of the nodes are randomly selected and masked as test nodes.
Each dataset is split chronologically, with the first 80\% for training and the last 20\% for testing.

\begin{table*}[t]
\centering
{
\begin{tabular}{lcccccccc}
\toprule
\multirow{2}{*}{Methods} & \multicolumn{4}{c}{\textbf{MeteoNet}} & \multicolumn{4}{c}{\textbf{RAIN-F}} \\
\cmidrule(lr){2-5} \cmidrule(lr){6-9}
 & RMSE↓ & MAE↓ & NSE↑ & CC↑ & RMSE↓ & MAE↓ & NSE↑ & CC↑ \\
\midrule
\N~\texttt{w/o Radar}      & 0.757 & 0.096 & 0.126 & 0.358 & 0.599 & 0.085 & 0.769 & 0.878 \\
\N~\texttt{w/o AWS}       & 0.657 & 0.076 & 0.343 & 0.587 & 0.891 & 0.169 & 0.490 & 0.739 \\
\N~\texttt{w/o RFE}        & 0.660 & 0.074 & 0.337 & 0.581 & 0.335 & 0.028 & 0.928 & 0.963 \\
\N~\texttt{w/o BPA-MHA}       & 0.652 & 0.070 & 0.352 & 0.594 & 0.353 & 0.028 & 0.920 & 0.959 \\
\N~\texttt{w/o CSTA}      & 0.651 & 0.065 & 0.353 & 0.594 & 0.373 & 0.028 & 0.911 & 0.954 \\
\N & \textbf{0.647} & \textbf{0.070} & \textbf{0.363} & \textbf{0.605} & \textbf{0.320} & \textbf{0.025} & \textbf{0.934} & \textbf{0.967} \\
\bottomrule
\end{tabular}
}
\caption{Ablation study on the \textbf{MeteoNet} and \textbf{RAIN-F} datasets.}
\label{tab:ablation_combined}
\end{table*}

\subsection{Overall Performance}
Table~\ref{tab:overall} compares \N\ with representative models from three categories in MeteoNet and RAIN-F.

\begin{itemize}[leftmargin=*]
    \item \N\ consistently outperforms all baselines on both MeteoNet and RAIN-F, reduced RMSE by 9.28\% and 47.37\%, MAE by 13.31\% and 70.71\%, while boosting NSE by 60.75\% and 22.56\%, and CC by 11.15\% and 10.48\%, respectively.
    These substantial gains highlight \N's ability to accurately recover fine-grained rainfall fields by fusing radar-informed inductive biases with structure-aware, multi-resolution modeling.

    \item Statistical baselines (\textbf{TPS, TIN}) assume spatial continuity, yielding good results under smooth rainfall conditions—particularly on RAIN-F, which is pre-interpolated.
    Thus, they struggle to capture sharp structural variations and localized extremes, leading to poor performance on MeteoNet, where AWS stations are sparse.

    \item Radar-based baselines (\textbf{Z-R, QPENet}) perform relatively well on MeteoNet, second only to \N, likely due to the stable precipitation patterns in France's temperate oceanic climates.
    In contrast, their performance drops significantly on RAIN-F,
    where interpolated AWS data are mismatched with raw radar observations.

    \item Deep learning baselines (\textbf{SSIN, GSI, STFNN}) show suboptimal performance across both datasets.
    On RAIN-F, interpolated and overly smooth AWS data weaken spatial variability, limiting the benefit of complex models.
    On MeteoNet, their coarse representations struggle to capture sharp local variations in raw, sparse gauge measurements.

\end{itemize}

\subsection{Key Component Assessment}




To assess the impact of each design in \N, we perform ablation experiments cross two datasets (Table~\ref{tab:ablation_combined}).
The results yield three key insights:

\begin{itemize}[leftmargin=*]

    \item \textbf{Radar provides spatial priors for convective structure.}
    Removing radar input (\N~\texttt{w/o Radar}) leads to substantial degradation—for instance, on MeteoNet (NSE $\downarrow$ 65.19\%, CC $\downarrow$ 40.53\%)—confirming radar's role in reconstructing fine-scale spatial patterns.

    \item \textbf{AWS observations correct radar ambiguity.}
    Removing AWS input (\N~\texttt{w/o AWS}) increases RMSE and reduces NSE, showing that sparse but trustworthy AWS measurements are essential for intensity calibration and reducing overestimation.

    \item We further evaluate the impact of structure-aware modules.
    First, removing the RainFront Encoder (\N~\texttt{w/o RFE}) degrades all metrics, highlighting the value of edge-aware spatial encoding.
    Second, replacing bidirectional fusion with radar-to-AWS-only alignment (\N~\texttt{w/o BPA-MHA}) weakens NSE and increases MAE, confirming the need of AWS-to-radar alignment.
    Finally, disabling geo-aware attention loss (\N~\texttt{w/o CSTA}) further reduces temporal correlation, showing that incorporating spatial proximity is key for modeling evolving rainfall dynamics.

\end{itemize}

\subsection{Sensitivity Analysis}
\begin{figure}[h!]
\centering
\includegraphics[width=1\linewidth]{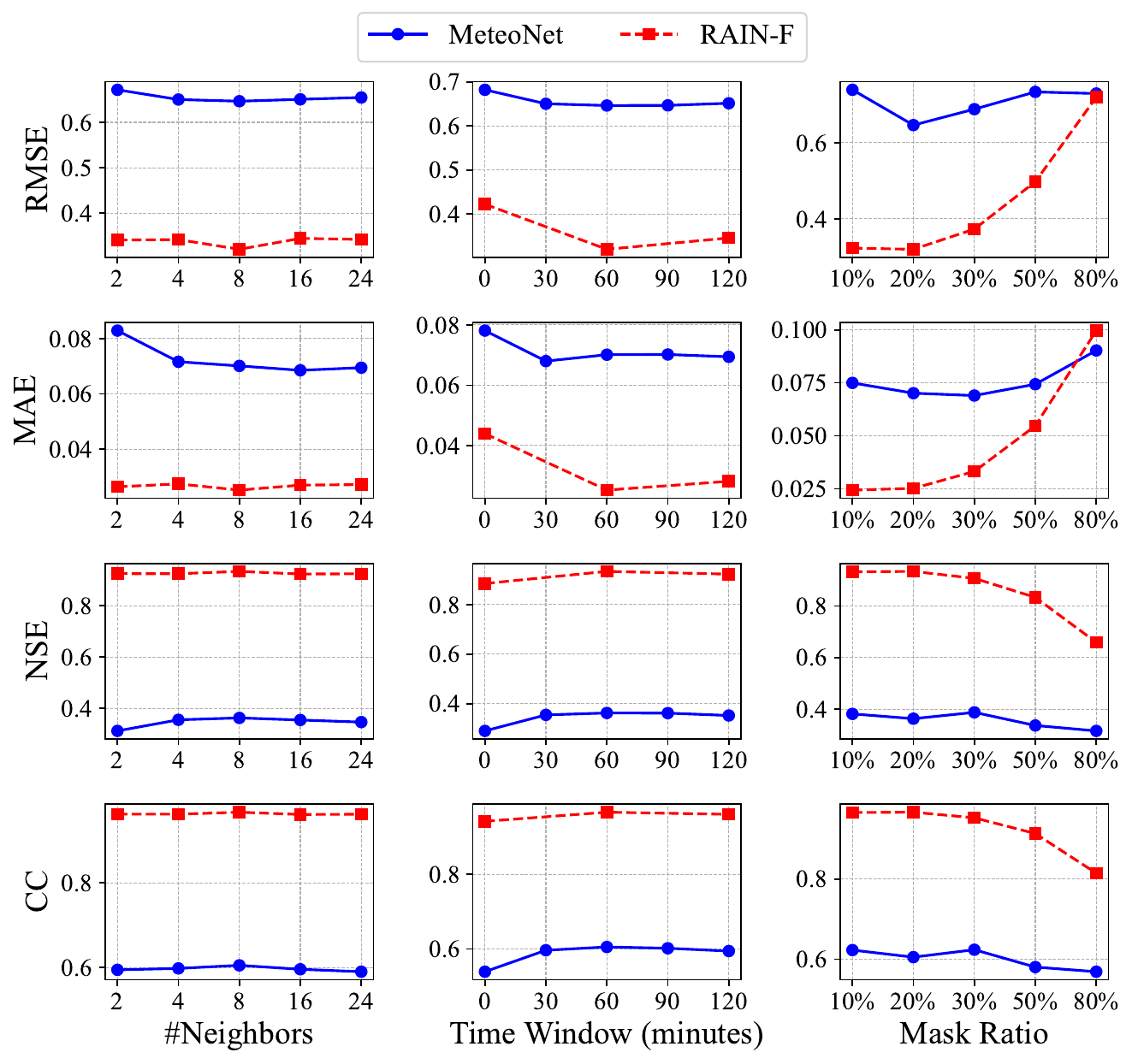}
\caption{
Performance(RMSE/MAE/NSE/CC) Comparison w.r.t. \#Neighbors, Time Window, and Mask Ratios.
}
\label{fig:senstivity}
\end{figure}

Figure~\ref{fig:senstivity} illustrates a sensitivity analysis on three key hyperparameters:
(i) the number of neighbors in AWS encoder,
(ii) the temporal length of the prior tracker,
and (iii) the mask ratio of AWS stratification.
Firstly, increasing AWS neighbors initially improves performance but declines beyond a threshold.
This reflects Tobler's first law of geography: nearby stations offer more relevant information, while distant ones introduce noise.
Secondly, extending the temporal window enhances performance as it captures persistent trends.
However, overly long histories may inject outdated or misleading signals due to the abrupt variability of rainfall fields.
Finally, moderate AWS masking improves generalization by encouraging the model to learn broader spatial dependencies.
Yet high mask ratios degrade performance by erasing critical local details needed for accurate inference.

\subsection{Case Study}

\begin{figure}[h]
\centering

\begin{subfigure}{1\linewidth}
    \centering
    \includegraphics[width=1\linewidth]{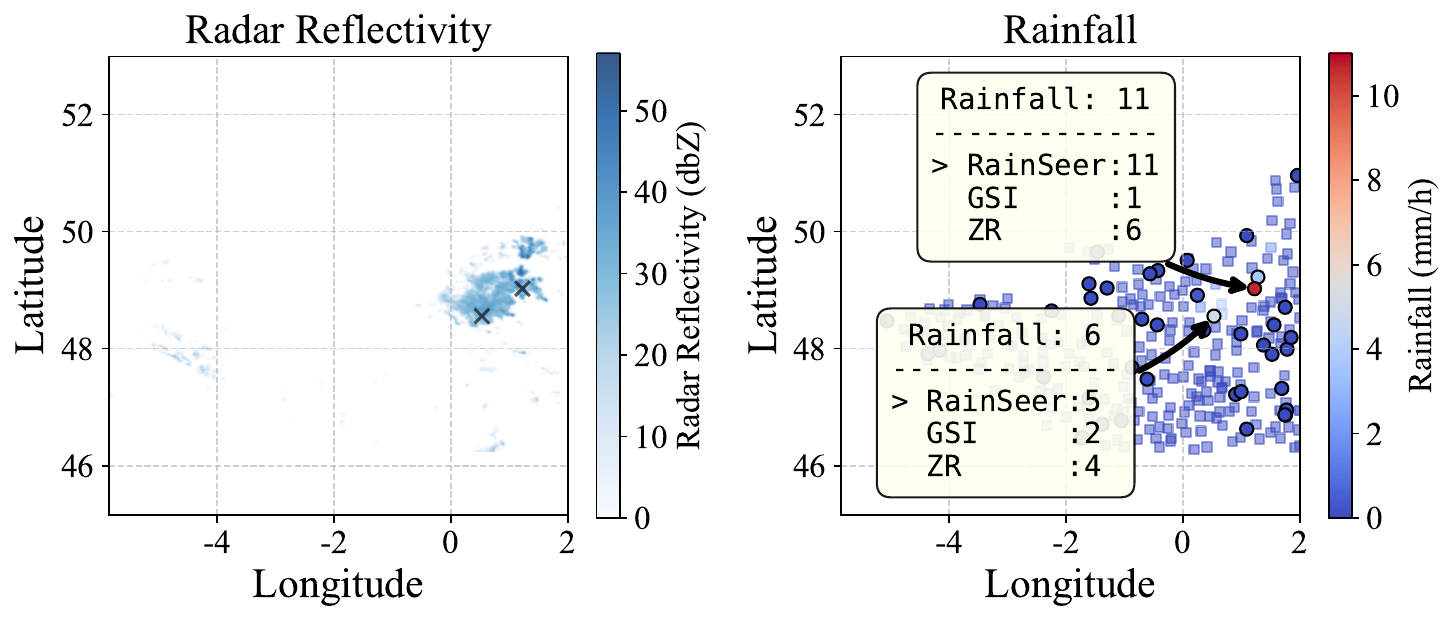}
    \caption{MeteoNet.}
\end{subfigure}

\begin{subfigure}{1\linewidth}
    \centering
    \includegraphics[width=1\linewidth]{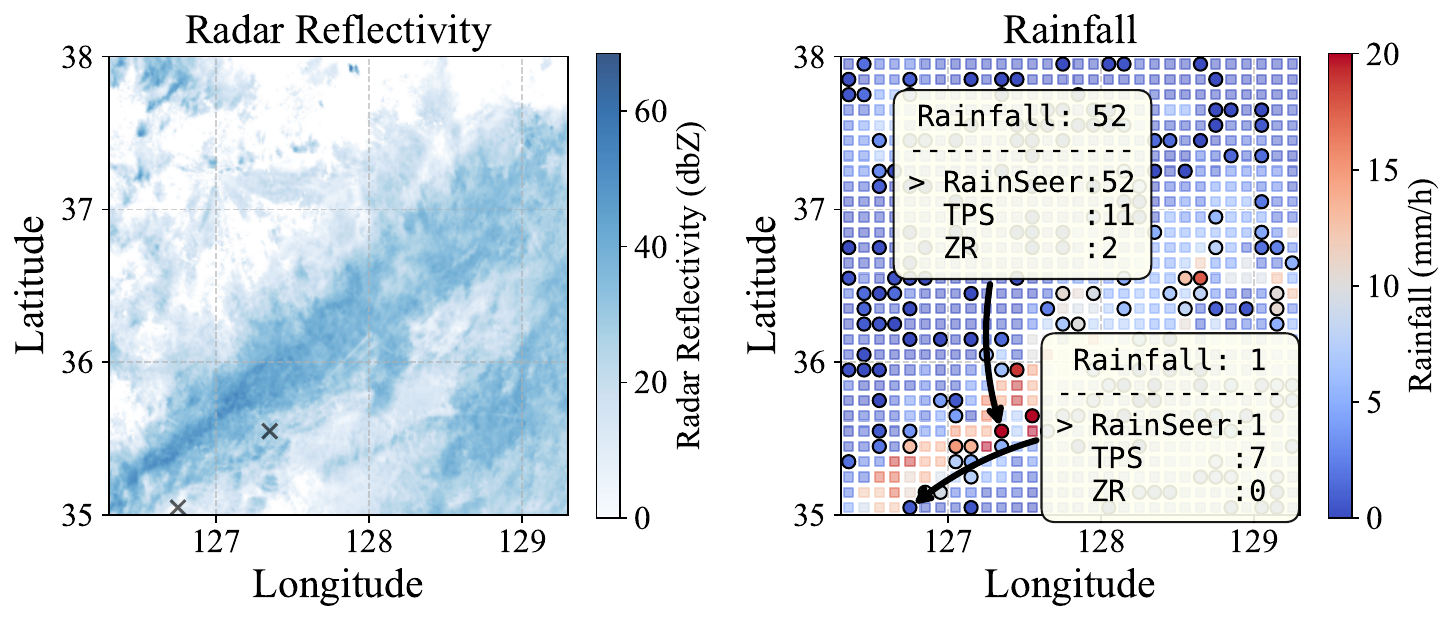}
    \caption{RAIN-F.}
\end{subfigure}

\caption{
    Visualization of Reconstruction Rainfall.
}
\label{fig:case}
\end{figure}


To evaluate~\N's robustness under abrupt rainfall transitions,
we present two case studies from distinct climates:
(a) temperate continental (MeteoNet), dominated by light rainfall ($z{<}20$ dBZ), and
(b) temperate monsoon (RAIN-F), characterized by intense convection ($z{>}50$ dBZ).
Figure~\ref{fig:case} visualizes predicted rainfall using a heatmap, where the origin indicates the reconstructed field and squares mark surface AWS observations. It highlights that baseline models often misplace or over-smooth key rainfall features—for instance, TPS significantly overestimates a low-rainfall region in RAIN-F  (GT = 1 mm/h, TPS = 7 mm/h).
In contrast,~\N\ provides predictions closely aligned with AWS observations, demonstrating strong generalization across diverse precipitation regimes.

%% file: 7conclusion.tex
\section{Conclusion}
\label{sec:conclusion}

In this study, we introduce~\N, a structure-aware framework for reconstructing fine-grained rainfall fields from heterogeneous meteorological signals.
Rather than treating radar reflectivity as auxiliary input,~\N\ reframes it as a physically grounded, dynamically evolving prior—capturing both mesoscale structure and physical process. 
To address the resolution and semantic gaps between aloft radar signals and surface rainfall,~\N\ designs two technical components:  
(i) \ComponentA\ projects volumetric radar structures onto point-wise rainfall via a \textit{bidirectional, spatially-constrained mapping};  
and (ii) \ComponentB\ models how hydro-meteors transform during descent—through melting and evaporation—using a \textit{causal spatiotemporal attention mechanism} that enforces physical plausibility and temporal coherence.
Extensive experiments on two real-world datasets demonstrate that~\N\ outperforms state-of-the-art baselines, reducing MAE by over 13.31\%.
Overall, \N\ lays a critical foundation for accurate flood forecasting, rapid disaster response, and resilient environmental decision-making.

%% file: arxiv-submission.bbl
\begin{thebibliography}{36}
\providecommand{\natexlab}[1]{#1}

\bibitem[{Abdollahipour, Ahmadi, and Aminnejad(2022)}]{abdollahipour2022review}
Abdollahipour, A.; Ahmadi, H.; and Aminnejad, B. 2022.
\newblock A review of downscaling methods of satellite-based precipitation estimates.
\newblock \emph{Earth Science Informatics}, 15(1): 1--20.

\bibitem[{Clevert, Unterthiner, and Hochreiter(2016)}]{clevert2016fastaccuratedeepnetwork}
Clevert, D.-A.; Unterthiner, T.; and Hochreiter, S. 2016.
\newblock Fast and Accurate Deep Network Learning by Exponential Linear Units (ELUs).
\newblock arXiv:1511.07289.

\bibitem[{Devia, Ganasri, and Dwarakish(2015)}]{devia2015review}
Devia, G.~K.; Ganasri, B.~P.; and Dwarakish, G.~S. 2015.
\newblock A review on hydrological models.
\newblock \emph{Aquatic procedia}, 4: 1001--1007.

\bibitem[{Feng et~al.(2024)Feng, Wang, Xia, Huang, Zhong, and Liang}]{ijcai2024p803}
Feng, Y.; Wang, Q.; Xia, Y.; Huang, J.; Zhong, S.; and Liang, Y. 2024.
\newblock Spatio-Temporal Field Neural Networks for Air Quality Inference.
\newblock In Larson, K., ed., \emph{Proceedings of the Thirty-Third International Joint Conference on Artificial Intelligence, {IJCAI-24}}, 7260--7268. International Joint Conferences on Artificial Intelligence Organization.
\newblock AI for Good.

\bibitem[{Guo et~al.(2019)Guo, Lin, Feng, Song, and Wan}]{guo2019attention}
Guo, S.; Lin, Y.; Feng, N.; Song, C.; and Wan, H. 2019.
\newblock Attention based spatial-temporal graph convolutional networks for traffic flow forecasting.
\newblock In \emph{Proceedings of the AAAI conference on artificial intelligence}, volume~33, 922--929.

\bibitem[{Hrachowitz and Weiler(2011)}]{hrachowitz2011uncertainty}
Hrachowitz, M.; and Weiler, M. 2011.
\newblock Uncertainty of precipitation estimates caused by sparse gauging networks in a small, mountainous watershed.
\newblock \emph{Journal of Hydrologic Engineering}, 16(5): 460--471.

\bibitem[{Hutchinson(1995)}]{hutchinson1995interpolating}
Hutchinson, M.~F. 1995.
\newblock Interpolating mean rainfall using thin plate smoothing splines.
\newblock \emph{International journal of geographical information systems}, 9(4): 385--403.

\bibitem[{Li et~al.(2023{\natexlab{a}})Li, Shen, Chen, and Ng}]{li2023rainfall}
Li, J.; Shen, Y.; Chen, L.; and Ng, C. W.~W. 2023{\natexlab{a}}.
\newblock Rainfall spatial interpolation with graph neural networks.
\newblock In \emph{International conference on database systems for advanced applications}, 175--191. Springer.

\bibitem[{Li et~al.(2023{\natexlab{b}})Li, Shen, Chen, and Ng}]{li2023ssin}
Li, J.; Shen, Y.; Chen, L.; and Ng, C. W.~W. 2023{\natexlab{b}}.
\newblock SSIN: Self-supervised learning for rainfall spatial interpolation.
\newblock \emph{Proceedings of the ACM on Management of Data}, 1(2): 1--21.

\bibitem[{Li, Chen, and Han(2023)}]{li2023polarimetric}
Li, W.; Chen, H.; and Han, L. 2023.
\newblock Polarimetric radar quantitative precipitation estimation using deep convolutional neural networks.
\newblock \emph{IEEE Transactions on Geoscience and Remote Sensing}, 61: 1--11.

\bibitem[{Liang et~al.(2023)Liang, Xia, Ke, Wang, Wen, Zhang, Zheng, and Zimmermann}]{liang2023airformer}
Liang, Y.; Xia, Y.; Ke, S.; Wang, Y.; Wen, Q.; Zhang, J.; Zheng, Y.; and Zimmermann, R. 2023.
\newblock Airformer: Predicting nationwide air quality in china with transformers.
\newblock In \emph{Proceedings of the AAAI conference on artificial intelligence}, volume~37, 14329--14337.

\bibitem[{Lu and Wong(2008)}]{lu2008adaptive}
Lu, G.~Y.; and Wong, D.~W. 2008.
\newblock An adaptive inverse-distance weighting spatial interpolation technique.
\newblock \emph{Computers \& geosciences}, 34(9): 1044--1055.

\bibitem[{Lucas et~al.(2022)Lucas, Longman, Giambelluca, Frazier, Mclean, Cleveland, Huang, and Lee}]{lucas2022optimizing}
Lucas, M.~P.; Longman, R.~J.; Giambelluca, T.~W.; Frazier, A.~G.; Mclean, J.; Cleveland, S.~B.; Huang, Y.-F.; and Lee, J. 2022.
\newblock Optimizing automated kriging to improve spatial interpolation of monthly rainfall over complex terrain.
\newblock \emph{Journal of Hydrometeorology}, 23(4): 561--572.

\bibitem[{Ly, Charles, and Degr{\'e}(2013)}]{ly2013different}
Ly, S.; Charles, C.; and Degr{\'e}, A. 2013.
\newblock Different methods for spatial interpolation of rainfall data for operational hydrology and hydrological modeling at watershed scale: a review.
\newblock \emph{Biotechnologie, agronomie, soci{\'e}t{\'e} et environnement}, 17(2).

\bibitem[{MeteoNet(2021)}]{MeteoNet}
MeteoNet. 2021.
\newblock MeteoNet.
\newblock \url{https://meteofrance.github.io/meteonet/}.
\newblock Accessed: 2025-06-20.

\bibitem[{Miller(2004)}]{miller2004tobler}
Miller, H.~J. 2004.
\newblock Tobler's first law and spatial analysis.
\newblock \emph{Annals of the association of American geographers}, 94(2): 284--289.

\bibitem[{Pearl(2009{\natexlab{a}})}]{pearl2009causal}
Pearl, J. 2009{\natexlab{a}}.
\newblock Causal inference in statistics: An overview.

\bibitem[{Pearl(2009{\natexlab{b}})}]{pearl2009causality}
Pearl, J. 2009{\natexlab{b}}.
\newblock \emph{Causality}.
\newblock Cambridge university press.

\bibitem[{Peng et~al.(2022)Peng, Bao, Yang, Wei, Zhu, Qiao, Wang, and Li}]{peng2022radar}
Peng, W.; Bao, S.; Yang, K.; Wei, J.; Zhu, X.; Qiao, Z.; Wang, Y.; and Li, Q. 2022.
\newblock Radar Quantitative Precipitation Estimation Algorithm Based on Precipitation Classification and Dynamical ZR Relationship.
\newblock \emph{Water}, 14(21): 3436.

\bibitem[{Prein et~al.(2017)Prein, Liu, Ikeda, Trier, Rasmussen, Holland, and Clark}]{prein2017increased}
Prein, A.~F.; Liu, C.; Ikeda, K.; Trier, S.~B.; Rasmussen, R.~M.; Holland, G.~J.; and Clark, M.~P. 2017.
\newblock Increased rainfall volume from future convective storms in the US.
\newblock \emph{Nature Climate Change}, 7(12): 880--884.

\bibitem[{RAIN-F(2021)}]{RAIN-F}
RAIN-F. 2021.
\newblock RAIN-F.
\newblock \url{https://dataon.kisti.re.kr/search/view.do}.
\newblock Accessed: 2025-06-20.

\bibitem[{Sokol et~al.(2021)Sokol, Szturc, Orellana-Alvear, Popova, Jurczyk, and C{\'e}lleri}]{sokol2021role}
Sokol, Z.; Szturc, J.; Orellana-Alvear, J.; Popova, J.; Jurczyk, A.; and C{\'e}lleri, R. 2021.
\newblock The role of weather radar in rainfall estimation and its application in meteorological and hydrological modelling—A review.
\newblock \emph{Remote Sensing}, 13(3): 351.

\bibitem[{Song et~al.(2021)Song, Fellegara, Iuricich, and De~Floriani}]{song2021efficient}
Song, Y.; Fellegara, R.; Iuricich, F.; and De~Floriani, L. 2021.
\newblock Efficient topology-aware simplification of large triangulated terrains.
\newblock In \emph{Proceedings of the 29th International Conference on Advances in Geographic Information Systems}, 576--587.

\bibitem[{Szegedy et~al.(2015)Szegedy, Liu, Jia, Sermanet, Reed, Anguelov, Erhan, Vanhoucke, and Rabinovich}]{Szegedy_2015_CVPR}
Szegedy, C.; Liu, W.; Jia, Y.; Sermanet, P.; Reed, S.; Anguelov, D.; Erhan, D.; Vanhoucke, V.; and Rabinovich, A. 2015.
\newblock Going Deeper With Convolutions.
\newblock In \emph{Proceedings of the IEEE Conference on Computer Vision and Pattern Recognition (CVPR)}.

\bibitem[{Tan et~al.(2021)Tan, Xie, Zuo, Xing, Liu, Xia, and Zhang}]{tan2021coupling}
Tan, J.; Xie, X.; Zuo, J.; Xing, X.; Liu, B.; Xia, Q.; and Zhang, Y. 2021.
\newblock Coupling random forest and inverse distance weighting to generate climate surfaces of precipitation and temperature with multiple-covariates.
\newblock \emph{Journal of Hydrology}, 598: 126270.

\bibitem[{Veličković et~al.(2018)Veličković, Cucurull, Casanova, Romero, Liò, and Bengio}]{GAT}
Veličković, P.; Cucurull, G.; Casanova, A.; Romero, A.; Liò, P.; and Bengio, Y. 2018.
\newblock Graph Attention Networks.
\newblock arXiv:1710.10903.

\bibitem[{Vimelia, Azis, and Purwani(2024)}]{vimelia2024prediction}
Vimelia, W.; Azis, C.~C.; and Purwani, S. 2024.
\newblock ‘Prediction of rainfall data in the DKI Jakarta area using cubic spline interpolation.
\newblock \emph{Reading Time}, 2024: 3--21.

\bibitem[{Wackernagel and Wackernagel(2003)}]{wackernagel2003ordinary}
Wackernagel, H.; and Wackernagel, H. 2003.
\newblock Ordinary kriging.
\newblock \emph{Multivariate geostatistics: an introduction with applications}, 79--88.

\bibitem[{Wang et~al.(2024)Wang, Wu, Duan, Zhang, Wang, Peng, Zheng, Liang, and Wang}]{wang2024nuwadynamics}
Wang, K.; Wu, H.; Duan, Y.; Zhang, G.; Wang, K.; Peng, X.; Zheng, Y.; Liang, Y.; and Wang, Y. 2024.
\newblock NuwaDynamics: Discovering and updating in causal spatio-temporal modeling.
\newblock In \emph{The Twelfth International Conference on Learning Representations}.

\bibitem[{Westra et~al.(2014)Westra, Fowler, Evans, Alexander, Berg, Johnson, Kendon, Lenderink, and Roberts}]{westra2014future}
Westra, S.; Fowler, H.~J.; Evans, J.~P.; Alexander, L.~V.; Berg, P.; Johnson, F.; Kendon, E.~J.; Lenderink, G.; and Roberts, N. 2014.
\newblock Future changes to the intensity and frequency of short-duration extreme rainfall.
\newblock \emph{Reviews of Geophysics}, 52(3): 522--555.

\bibitem[{WIKIPEDIA(2021)}]{Europeanfloods}
WIKIPEDIA. 2021.
\newblock 2021 European floods.
\newblock \url{https://en.wikipedia.org/wiki/2021_European_floods}.
\newblock Accessed: 2025-06-20.

\bibitem[{Wu et~al.(2024{\natexlab{a}})Wu, Chen, Wang, Peng, Sun, and Chen}]{ijcai2024p269}
Wu, B.; Chen, W.; Wang, W.; Peng, B.; Sun, L.; and Chen, L. 2024{\natexlab{a}}.
\newblock WeatherGNN: Exploiting Meteo- and Spatial-Dependencies for Local Numerical Weather Prediction Bias-Correction.
\newblock In Larson, K., ed., \emph{Proceedings of the Thirty-Third International Joint Conference on Artificial Intelligence, {IJCAI-24}}, 2433--2441. International Joint Conferences on Artificial Intelligence Organization.
\newblock Main Track.

\bibitem[{Wu, Verlinde, and Sun(2000)}]{wu2000dynamical}
Wu, B.; Verlinde, J.; and Sun, J. 2000.
\newblock Dynamical and microphysical retrievals from Doppler radar observations of a deep convective cloud.
\newblock \emph{Journal of the atmospheric sciences}, 57(2): 262--283.

\bibitem[{Wu et~al.(2024{\natexlab{b}})Wu, Liang, Xiong, Zhou, Huang, Wang, and Wang}]{wu2024earthfarsser}
Wu, H.; Liang, Y.; Xiong, W.; Zhou, Z.; Huang, W.; Wang, S.; and Wang, K. 2024{\natexlab{b}}.
\newblock Earthfarsser: Versatile spatio-temporal dynamical systems modeling in one model.
\newblock In \emph{Proceedings of the AAAI conference on artificial intelligence}, volume~38, 15906--15914.

\bibitem[{Xu et~al.(2022)Xu, Ma, Yan, and Peng}]{xu2022era5}
Xu, J.; Ma, Z.; Yan, S.; and Peng, J. 2022.
\newblock Do ERA5 and ERA5-land precipitation estimates outperform satellite-based precipitation products? A comprehensive comparison between state-of-the-art model-based and satellite-based precipitation products over mainland China.
\newblock \emph{Journal of Hydrology}, 605: 127353.

\bibitem[{Zhang et~al.(2021)Zhang, Bi, Liu, Chen, Zhang, Shen, Yang, Wang, Zhang, and Yao}]{zhang2021deep}
Zhang, Y.; Bi, S.; Liu, L.; Chen, H.; Zhang, Y.; Shen, P.; Yang, F.; Wang, Y.; Zhang, Y.; and Yao, S. 2021.
\newblock Deep learning for polarimetric radar quantitative precipitation estimation during landfalling typhoons in South China.
\newblock \emph{Remote Sensing}, 13(16): 3157.

\end{thebibliography}
